\documentclass[letterpaper, 10 pt, conference]{ieeeconf}  

\IEEEoverridecommandlockouts                              

\usepackage{graphicx} 
\usepackage{subcaption}
\usepackage{balance}
\usepackage{mathptmx} 
\usepackage{times} 
\usepackage{amsmath} 
\usepackage{amssymb}  
\usepackage{cite}
\usepackage{caption}
\usepackage{xcolor}
\usepackage{multirow}
\usepackage{siunitx}
\usepackage{hyperref}
\usepackage{algorithm}
\usepackage{algpseudocode}
\usepackage{placeins}
\usepackage{booktabs}
\usepackage[super]{nth}
\usepackage[colorinlistoftodos]{todonotes}

\title{\LARGE \bf
Hardware-Accelerated Instance Segmentation for Resource-Constrained Space Robotics with Criticality Analysis
}
\author{
Siddhant Shete$^{1}$, Hilmi Dogu Kücüker$^{1}$, Udo Frese$^{1,2}$, Frank Kirchner$^{1,2}$%
\thanks{$^{1}$All authors are with the Robotics Innovation Center of the German Research Center for Artificial Intelligence GmbH (DFKI), Bremen, Germany.}%
\thanks{$^{2}$Frank Kirchner and Udo Frese are additionally affiliated with the University of Bremen, Germany.}%
}

\begin{document}

\maketitle
\thispagestyle{empty}
\pagestyle{empty}

\begin{abstract}

Autonomous lunar missions require real-time perception under three coupled constraints: extreme low-light conditions, limited onboard compute, and radiation-induced hardware faults that can silently corrupt inference.
We present a deployment-oriented instance segmentation framework for resource-constrained lunar robotics that jointly addresses quantization calibration and system-level fault exposure under strict compute constraints. First, we introduce Activation Variance Informative Sampling (AVIS), a label-free calibration strategy that deterministically selects calibration samples based on activation variance statistics.
Second, we deploy a YOLO-based segmentation model on a Deep Learning Processor Unit (DPU) with architectural modifications that reduce CPU fallback paths and enable statically compiled execution with bounded latency in low-lighting conditions.
We further introduce a software-level criticality analysis to estimate fault exposure and guide mitigation under radiation-constrained operation.
On a lunar micro-rover platform, AVIS with bias correction recovers 69.8\% of quantization-induced accuracy loss while achieving 309 ms inference latency and 5.7 W power consumption. Targeted mitigation reduces global criticality by 31.7\%.
The results demonstrate an integrated approach and a blueprint for a reliable and safe AI perception framework under space deployment constraints.

\end{abstract}


\section{Introduction}
\label{sec:introduction}

Autonomous lunar robots must perceive their environment with high reliability
to detect hazards, plan safe trajectories, and execute mission-critical tasks
without human intervention~\cite{Amaya_Mej_a_2024}. Communication latencies
of several seconds, harsh environmental extremes, and the complete absence of
in-situ repair make real-time onboard autonomy a prerequisite for mission
success~\cite{AutonomousAIinSpaceExploration}. However, standard AI models are unsuitable for the lunar operating envelope, where three linked limitations—extreme low-light illumination, extremely limited onboard computing, and radiation-induced hardware defects—can decrease both perception accuracy and runtime predictability.

Perception failures in autonomous systems can have mission-critical consequences.
The Chandrayaan-2 Vikram lander crash, which has been widely analyzed in the literature as highlighting the importance of real-time fault handling during powered descent~\cite{Chandrayaan-2}, resulted
in complete mission loss; the engineering lessons directly informed the
successful Chandrayaan-3 redesign~\cite{Chandrayaan-3}. These missions highlight the need for deterministic execution, autonomous fault tolerance, and graceful degradation, because in-flight retraining, manual intervention, and hardware
repair are infeasible~\cite{SurveyonAIEnabledCV}. From a deployment perspective, three tightly coupled constraints dominate system design:
\begin{enumerate}
\item \textbf{Illumination:} Lunar scenes exhibit extreme low-light regions,
high-contrast cast shadows, and specular regolith reflections that confound
conventional visual perception~\cite{Illumination}.

\item \textbf{Compute constraints:} Onboard platforms are severely limited in
power dissipation, memory capacity, and thermal headroom~\cite{chen2018tvmautomatedendtoendoptimizing, Issues}, making computationally efficient deployment mandatory~\cite{ArtificialIntelligenceinResource-Constrained}.

\item \textbf{Radiation:} Single-event effects—transient bit flips, memory
corruption, and logic upsets—can silently corrupt inference without immediate observable indications~\cite{electronics14142896}.
\end{enumerate}

\noindent Consequently, flight-grade perception must be deterministic,
resilient to transient faults, and capable of self-recovery, prioritizing
mission safety over peak benchmark performance.

Deep neural networks achieve state-of-the-art segmentation accuracy on
terrestrial benchmarks, but their behavior degrades unpredictably under
space-deployment conditions: quantization noise from reduced precision,
activation distortion from model compression, and silent data corruption from
radiation~\cite{Pang2017RobustDL, Caroselli}. Post-training quantization (PTQ)
reduces model size and latency—prerequisites for embedded deployment—but introduces layer-dependent accuracy loss that is difficult to bound analytically~\cite{Zhou_Moosavi-Dezfooli_Cheung_Frossard_2018}. Instance segmentation provides the dense, geometry-aware scene understanding required for hazard avoidance: pixel-level delineation of rocks and craters supports landing site assessment, real-time obstacle detection, and autonomous target selection~\cite{s22218393, 10115717, Amaya_Mej_a_2024}. Yet the computational intensity of segmentation networks significantly complicates their deployment, particularly under strict power and memory constraints.

In this work, we present a deployment-oriented instance segmentation framework for lunar robotics that jointly addresses quantization fidelity, execution determinism, and radiation-related fault exposure. The framework integrates three components:

\begin{itemize}
\item a label-free calibration strategy (AVIS) that selects informative inputs using activation variance statistics to improve INT8 quantization stability,
\item a hardware-aware deployment pipeline that reduces runtime variability through statically compiled execution on a DPU accelerator, and
\item a software-level criticality analysis that ranks functional blocks by memory footprint and execution time to guide radiation-informed mitigation.
\end{itemize}

These components are designed to operate jointly under strict compute, power, and reliability constraints of lunar-class embedded robotics. AVIS reduces quantization sensitivity by improving calibration stability, while the criticality analysis identifies software components most exposed to radiation-induced faults. Together, they improve robustness under the combined constraints of lunar deployment.

Evaluated on a lunar micro-rover platform, the framework recovers 69.8\% of quantization-induced accuracy loss, achieves 309\,ms inference at 5.7\,W, and reduces radiation-induced criticality by 31.7\% through targeted mitigation. This demonstrates a deployable instance segmentation system for lunar robotics under combined constraints of quantization, deterministic execution, and radiation-induced faults.


\section{Related Work}
\label{sec:related_work}

We situate our contribution at the intersection of planetary perception,
post-training quantization for edge deployment, and radiation-tolerant AI
execution.

\paragraph{Planetary Perception}
Deep learning-based perception is increasingly explored for planetary
navigation~\cite{VisualNavigation}, hazard avoidance~\cite{engproc2025090044},
and terrain segmentation from orbital and surface
imagery~\cite{LunarGroundSegmentation, TerrainSemantic,
bouldersInstanceSegmentation}. However, these approaches are evaluated
exclusively on offline datasets with no consideration of deterministic
execution, bounded latency, or deployment on radiation-informed hardware. The
implicit assumption of unlimited floating-point compute and safe execution
environments limits their applicability to flight-grade systems.

\paragraph{Post-Training Quantization and Calibration}
PTQ and model compression enable deployment on space-grade edge hardware
including FPGAs and DPUs~\cite{RapidPrototyping, Optimizing}. INT8 quantization
reduces memory footprint and latency, but its effectiveness depends
on calibration data selection~\cite{pmlr-v139-hubara21a}. Poorly chosen inputs
cause activation range misestimation, leading to clipping artifacts and
unpredictable accuracy degradation. Recent work compares quantization against
pruning~\cite{kuzmin2024pruningvsquantizationbetter} and explores label-free
calibration strategies~\cite{Williams_2024,
Zhou_Moosavi-Dezfooli_Cheung_Frossard_2018}, but existing methods do not
guarantee the deterministic, reproducible quantization behavior required by
safety-critical flight software.

This gap motivates AVIS, which provides deterministic sample selection without
labels or retraining. Unlike prior post-training quantization methods that rely
on activation range coverage, entropy-based sampling, or reconstruction loss
minimization, AVIS constructs a deterministic ranking of calibration samples
based on aggregated layer-wise activation variance. This enables reproducible
sample selection without stochastic sampling or label dependence.

\paragraph{Radiation-Tolerant AI Acceleration}
The space radiation environment induces single-event upsets that silently
corrupt weights, activations, or control logic in embedded
accelerators~\cite{electronics14071421, electronics14142896}. Hardware-level
mitigations—Triple Modular Redundancy (TMR), Error Detection and Correction (EDAC) codes, and memory scrubbing—significantly improve
resilience~\cite{turn0search3, turn0academia19, turn0academia21}, but address
radiation tolerance at the accelerator level in isolation. However, these methods primarily focus on hardware-level mitigation and largely
neglect software-level fault propagation across inference pipelines. This limits
efficient mitigation resource allocation under the severe power, area, and weight
budgets of space platforms~\cite{Energy-aware, GOODWILL2025239}. Recent studies
on fault injection in neural networks highlight sensitivity to bit-level
perturbations, but do not provide deployment-level criticality prioritization.

\paragraph{Gap and Positioning}
Table~\ref{tab:related_positioning} summarizes coverage across key deployment
dimensions. To the best of our knowledge, no prior work simultaneously integrates
deterministic label-free PTQ calibration, DPU-accelerated instance segmentation execution, and
function-wise radiation-informed criticality analysis within a unified perception
framework for space robotics.

\begin{table}[h]
\centering
\caption{Comparison of prior work across key deployment dimensions. Our framework integrates instance segmentation, post-training quantization, DPU deployment, radiation-informed analysis, and label-free calibration within a unified system.}
\label{tab:related_positioning}
\footnotesize
\setlength{\tabcolsep}{4pt}
\begin{tabular}{lccccc}
\toprule
\textbf{Work} 
& \textbf{Instance} 
& \textbf{PTQ} 
& \textbf{DPU} 
& \textbf{R.A} 
& \textbf{Label-Free} \\
& \textbf{Segmentation} 
& \textbf{} 
& 
& 
& \textbf{Calibration} \\
\midrule
Petrakis et al.~\cite{LunarGroundSegmentation}      & \checkmark & -- & -- & -- & -- \\
Liu et al.~\cite{TerrainSemantic}                 & \checkmark & -- & -- & -- & -- \\
Prieur et al.~\cite{bouldersInstanceSegmentation}   & \checkmark & -- & -- & -- & -- \\
Blacker et al.~\cite{RapidPrototyping}              & --         & \checkmark & -- & -- & -- \\
Williams et al.~\cite{Williams_2024}                & --         & \checkmark & -- & -- & \checkmark \\
Shao et al.~\cite{turn0search3}                     & --         & -- & \checkmark & \checkmark & -- \\
Tedeschi et al.~\cite{turn0academia19}              & --         & -- & -- & \checkmark & -- \\
\textbf{Ours}                                      & \checkmark & \checkmark & \checkmark & \checkmark & \checkmark \\
\bottomrule
\end{tabular}

\vspace{2pt}
\footnotesize \textbf{Note:} R.A = Radiation Analysis
\end{table}


\vspace{-15pt}
\section{System Design and Problem Formulation}
\label{sec:system_design}

The perception framework targets resource-constrained lunar surface platforms
where autonomous hazard detection must operate without ground-in-the-loop
support. The reference platform is the Lunar Micro Rover for Night Survival
(LuNiS)\footnote{\url{https://robotik.dfki-bremen.de/de/forschung/robotersysteme/lunis}}, a micro-class rover designed for
semi-autonomous navigation under extreme low-angle illumination near
permanently shadowed regions, with a continuous power budget below 10\,W for
all onboard computation. LuNiS Rover exemplifies systems where perception is
expected to be simultaneously accurate, deterministic, energy-efficient, and
resilient to radiation-induced faults.

\begin{figure*}[h]
    \centering
    \includegraphics[width=0.9\linewidth]{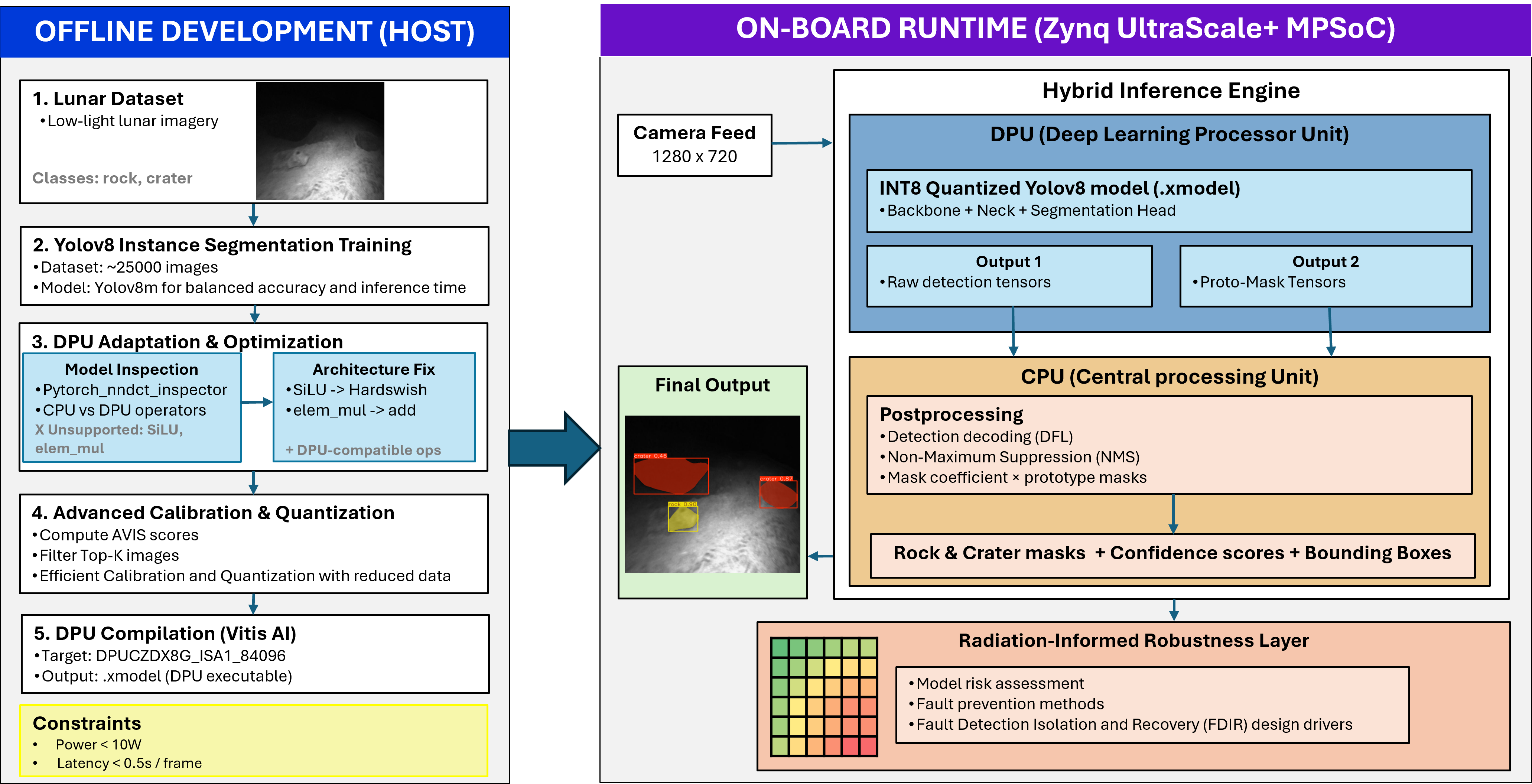}
    \caption{System architecture. \textbf{Left:} Offline
    phase—training, AVIS calibration selection, INT8 quantization
    with bias correction, and XMODEL (Vitis-AI model representation) compilation. \textbf{Right:}
    Onboard phase—DPU-accelerated convolution and deterministic
    CPU post-processing (non-maximum suppression (NMS), mask reconstruction). All weights are
    frozen; no in-flight modification is performed.}
    \label{fig:arch}
\end{figure*}

The deployable perception pipeline, illustrated in Fig.~\ref{fig:arch},
performs fully onboard instance segmentation of rocks and
craters—the primary geometric hazards during surface traversal. All model
weights are frozen after offline training; no in-flight retraining, fine-tuning,
or dynamic model modification is performed.

\subsection{Problem Formulation}
\label{sec:problem_formulation}

The objective is to deploy a high-accuracy instance segmentation model on a
resource-constrained DPU such that it is designed to operate deterministically under fixed execution graphs, bounded in
latency, and robust to both quantization-induced accuracy loss and
radiation-induced hardware faults. The deployment is designed to address the following requirements:

\begin{enumerate}

\item \textbf{Quantization fidelity:} Segmentation quality must be preserved
under INT8 post-training quantization with bounded, characterizable accuracy
degradation relative to the FP32 baseline. Non-reproducible accuracy loss is
undesirable for reliable deployment.

\item \textbf{Deterministic real-time execution:} End-to-end inference is
expected to complete within a fixed latency bound dictated by the rover's motion
profile. Identical inputs are expected to produce identical outputs under fixed
deployment conditions and static execution graphs, with no dependence on dynamic
resource allocation.

\item \textbf{Data-efficient, label-free calibration:} INT8 calibration must
operate without labeled data and select samples deterministically to ensure
repeatable quantized behavior, reflecting the scarcity of representative
lunar imagery and the impossibility of post-launch adjustment.

\item \textbf{Radiation-aware fault robustness:} The framework estimates
per-block vulnerability to radiation-induced single-event effects and supports
allocation of mitigation resources to the highest-risk components.

\end{enumerate}

\noindent Quantization actions impact radiation-induced bit flip sensitivity, while hardware partitioning between CPU and DPU affects latency and spatial radiation exposure. Calibration strategy additionally influences INT8 dynamic range tightness, affecting fidelity and robustness. This connection drives the co-design approach outlined in Section~\ref{sec:methodology}.


\section{Methodology}
\label{sec:methodology}

This section presents the four methodological components of our framework,
designed to jointly satisfy the deployment requirements defined in
Section~\ref{sec:problem_formulation}: segmentation architecture
(Section~\ref{sec:seg}), hardware-deterministic deployment
(Section~\ref{sec:hw_adapt}), activation-variance-based calibration
(Section~\ref{sec:avis}), and radiation-aware criticality analysis
(Section~\ref{sec:criticality}).

\subsection{Instance Segmentation Architecture}
\label{sec:seg}

We employ YOLOv8m-based instance segmentation to detect and delineate rocks and
craters—the primary geometric hazards during surface traversal~\cite{s22218393, 10115717}. Instance segmentation is preferred over
bounding-box detection because pixel-level boundary delineation is essential for
trajectory planning over irregular terrain, where box approximations introduce
unacceptable collision-risk margins. The medium variant was selected for its trade-off between mask accuracy and
inference efficiency: larger variants exceed DPU resource limits, while smaller
variants exhibit measurable degradation on thin or partially occluded rock
boundaries. The model is trained offline on 25{,}000 images comprising synthetic
renders and custom-collected data~\cite{shete_2025_14673237} under
lunar-representative conditions (low-light illumination, high-contrast shadows,
specular regolith reflections). All weights are frozen after training, ensuring deterministic and reproducible inference throughout the mission lifetime.

\subsection{Hardware-Aware Model Adaptation for DPU Deployment}
\label{sec:hw_adapt}

Direct deployment of YOLOv8m on the target DPU is not feasible without
architectural adaptation. The Vitis~AI toolchain does not natively support
segmentation network compilation, and several architectural components produce
compilation failures or CPU fallbacks that violate latency bounds and introduce
non-deterministic execution. The deployment targets a Zynq UltraScale+ MPSoC
with the \texttt{DPUCZDX8G\_ISA1\_B4096} DPU. Three categories of adaptation
achieve full-graph compilation:

\paragraph{Activation function replacement}
YOLOv8's SiLU and Swish activations require element-wise sigmoid, which is unsupported or
poorly quantized on the target DPU. We replace all activations with HardSwish,
which is natively supported and preserves nonlinear expressiveness~\cite{howard2017mobilenetsefficientconvolutionalneural}.
Alternatives were systematically evaluated: ReLU variants introduced excessive
sparsity degrading mask completeness, while HardSigmoid produced asymmetric
quantization noise in deeper layers.

\paragraph{Elimination of unsupported operations}
The attention mechanism's scaled dot-product computation employs a DPU-unsupported operator (\texttt{nndct\_elemwise\_mul}), resulting in CPU fallback. We redesign this as a fully DPU-compatible additive projection that preserves cross-channel feature reweighting using only supported operations~\cite{cai2021graphnormprincipledapproachaccelerating}.

\paragraph{Static tensor dimensions}
Dynamic shape-dependent operations (\texttt{nndct\_shape},
\texttt{aten::meshgrid}, \texttt{nndct\_arange}) default to CPU execution. We
replace runtime shape queries with fixed dimensions derived from the network's
multi-scale feature pyramid structure, enabling most operations to execute on the DPU under the target compilation constraints and ensuring bounded execution latency.

\noindent Together, these adaptations enable full-graph XMODEL compilation,
eliminate dynamic execution paths, and support highly predictable and reproducible inference latency under the target toolchain constraints.

\subsection{Activation Variance Informative Sampling (AVIS)}
\label{sec:avis}

PTQ accuracy is sensitive to calibration data selection: non-representative
inputs may cause activation saturation or dynamic range under-utilization, leading to
unpredictable accuracy degradation~\cite{pmlr-v139-hubara21a,
Zhou_Moosavi-Dezfooli_Cheung_Frossard_2018}. We propose AVIS, a deterministic,
label-free strategy that selects highly informative inputs for calibration stability by
evaluating activation variance across network layers. The core hypothesis is that inputs producing higher spatial variance in intermediate feature maps tend to activate a broader subset of learned representations, which empirically improves calibration stability and INT8 scale estimation.

\paragraph{Activation variance scoring}
Let $\mathcal{X} = \{x_i\}_{i=1}^{N}$ denote the candidate pool and $L$ the
number of layers. For each candidate $x_i$:
\begin{equation}
s_i = \frac{1}{L} \sum_{l=1}^{L} \mathrm{mean}_{c}
\Big( \mathrm{Var}_{H,W} \big( A_l^{c}(x_i) \big) \Big)
\end{equation}

where $A_l^{c}(x_i) \in \mathbb{R}^{H \times W}$ is the spatial activation map.

Higher $s_i$ is interpreted as indicating inputs that activate a broader subset of learned representations and are therefore more informative for quantization scale estimation.

\paragraph{Deterministic Top-$K$ selection}
Calibration images are selected via:
\begin{equation}
\mathcal{X}_{\text{AVIS}} =
\left\{ x_i \;\middle|\; s_i \in \operatorname{TopK}\big(\{s_j \mid s_j > 0\},\, K\big) \right\}
\end{equation}

The positivity constraint excludes near-uniform or underexposed inputs. The resulting set is fully determined by the frozen network's activation statistics, with no dependence on labels or stochastic sampling, ensuring deterministic calibration under fixed model parameters. AVIS reduces calibration data requirements by approximately $2\times$ relative to random sampling while improving INT8 accuracy, and operates independently of training or architectural modifications.

\subsection{Quantized Deployment and CPU--DPU Partitioning}
\label{sec:deployment}

Following AVIS calibration, the network undergoes INT8 PTQ with bias correction, compensating for systematic layer-wise activation shifts that degrade mask boundary precision~\cite{nagel2019datafreequantizationweightequalization}. The resulting model is compiled into an XMODEL for DPU deployment. At runtime, the pipeline executes as a structured CPU--DPU hybrid:

\begin{itemize}
\item \textbf{DPU:} Backbone, neck, and segmentation head compiled as a single INT8 XMODEL graph with bounded latency.
\item \textbf{CPU:} Deterministic post-processing (confidence decoding, NMS, mask reconstruction) with statically defined memory and no dynamic control flow or heap allocation.
\end{itemize}

This partitioning ensures deterministic, bounded-latency execution at system level by eliminating CPU fallbacks and uncontrolled execution paths. It also enables structured isolation of compute-intensive kernels for downstream radiation-aware analysis.

\subsection{Radiation-Aware Criticality Analysis}
\label{sec:criticality}

This section quantifies vulnerability of each functional block to radiation-induced single-event effects (SEE), enabling targeted allocation of mitigation resources.

\subsubsection{Scope and Assumptions}

The analysis evaluates SEE susceptibility of memory-resident components (RAM and associated runtime buffers) under onboard execution. Mission-level particle flux and device cross-sections are not explicitly modeled. Flash memory effects are neglected due to demonstrated FPGA robustness~\cite{microchip_see_fpga_2011}. We focus on the functional impact of SEE at the software level. The analysis provides a relative criticality ranking rather than absolute failure probability.

\subsubsection{Occurrence Model}

Memory-level SEE events are approximated as spatially uniform at the memory layout level. The short inference cycle (1.2\,s) relative to mission duration supports a time-invariant approximation within each execution window:

\begin{equation}
p(x) = \frac{1}{\mathrm{Area}(R_M \cap R_T)}\,
\mathbf{1}_{R_M \cap R_T}(x)
\end{equation}

where $R_M$ and $R_T$ denote memory and temporal execution regions. This expression follows from modeling memory residency and execution time as independent uniform distributions over $R_M$ and $R_T$, respectively, which reduces to a uniform distribution over their intersection.

\subsubsection{Functional Block Decomposition}

We aggregate operations at the library level into functional blocks:

\begin{itemize}
\item \textbf{CPU-static:} Persistent allocations (OpenCV, runtime, math libraries), profiled via \textit{pmap}.
\item \textbf{CPU-dynamic:} Heap allocations during post-processing, profiled via \textit{massif}.
\item \textbf{DPU:} XMODEL kernel footprint and execution time, profiled via \textit{vaitrace}.
\end{itemize}

Temporal proportions are extracted via \textit{callgrind}.

\subsubsection{Criticality Classification}
\begin{figure}[h]
    \centering
    \includegraphics[width=0.8\columnwidth]{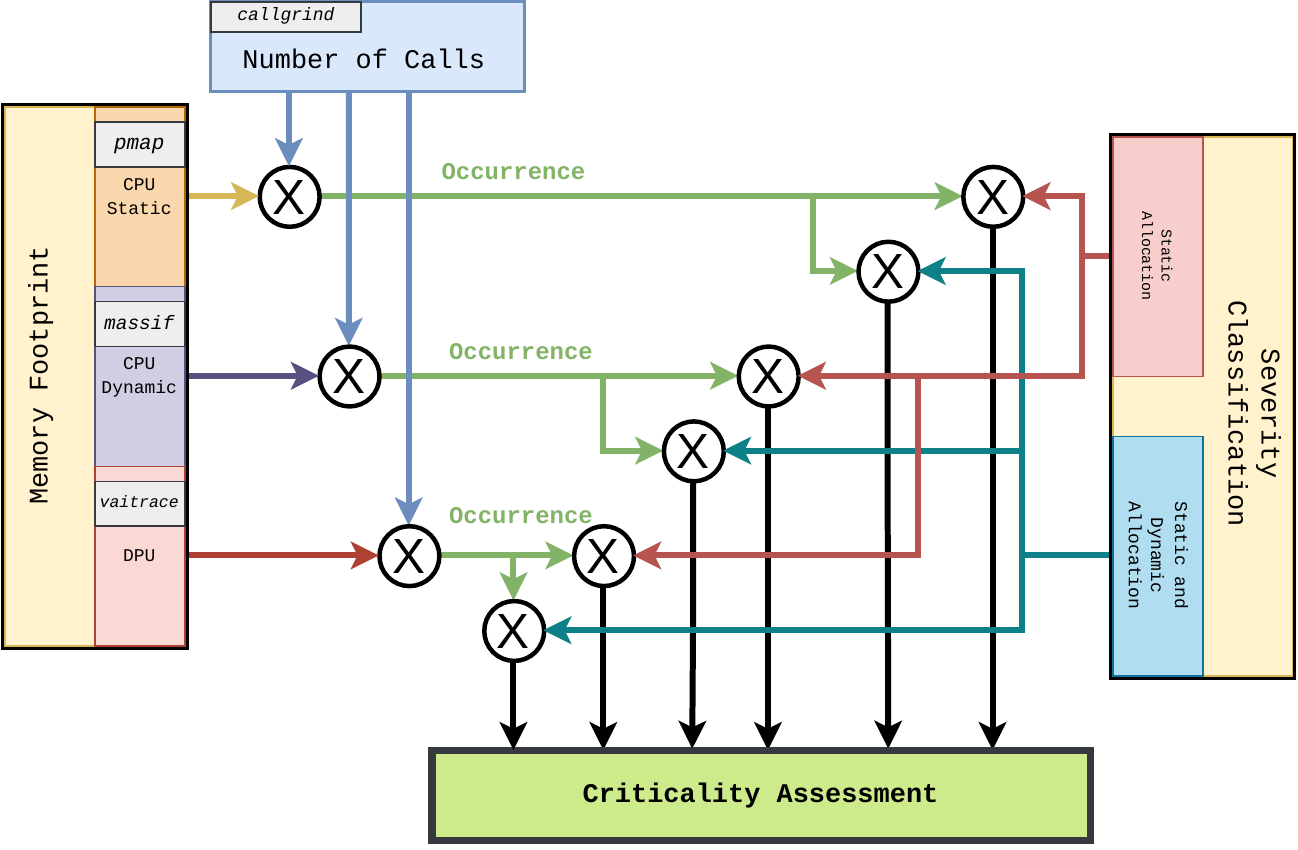}
    \caption{Risk assessment pipeline following ECSS methodology.}
    \label{fig:risk-analysis-pipe}
\end{figure}
Criticality follows ECSS methodology~\cite{ECSS-Q-ST-30-02C} as shown in Fig.~\ref{fig:risk-analysis-pipe}. Occurrence is derived from measured memory footprint and execution time. Severity reflects functional impact under fault conditions and is assigned based on detectability and propagation potential. Dynamic heap allocations are conservatively assigned high severity due to susceptibility to silent corruption and lack of deterministic recovery mechanisms. After mitigation, severity is reassigned based on software stack hierarchy, where lower-level libraries exhibit higher propagation impact.

The final criticality metric is defined as follows:

\begin{equation}
C = O \times S
\end{equation}
This formulation provides a first-order risk approximation consistent with standard FMEA practice rather than a physical failure probability model.


\section{Experimental Results}
\label{sec:experiments}

We evaluate the framework across three dimensions: segmentation accuracy under
quantization (Section~\ref{sec:seg_perf}), deployment efficiency
(Section~\ref{sec:inf_eff}), and radiation-informed robustness
(Section~\ref{sec:crit_results}).

\subsection{Segmentation Performance}
\label{sec:seg_perf}

\begin{table}[t]
\centering
\caption{Segmentation accuracy across platforms and quantization configurations.
GPU FP32 serves as the reference baseline (0\% loss). All INT8 models run on the hybrid CPU--DPU platform.}
\label{tab:seg_accuracy}
\footnotesize
\setlength{\tabcolsep}{3pt}   
\renewcommand{\arraystretch}{1.05}
\begin{tabular}{llcc}
\toprule
\textbf{Platform} & \textbf{Model} & \textbf{mAP / IoU} & \textbf{mAP drop} \\
\midrule
GPU     & Y8m FP32                     & 0.802 / 0.782 & 0.0\% \\
CPU-DPU & Y8m INT8 (RC)                & 0.749 / 0.731 & -6.6\% \\
CPU-DPU & Y8m INT8 + Bias Corr.\ (RC)  & 0.768 / 0.741 & -4.2\% \\
CPU-DPU & Y8m INT8 + AVIS + Bias Corr. & \textbf{0.786 / 0.767} & \textbf{-2.0\%} \\
\bottomrule
\end{tabular}

\vspace{2pt}
\footnotesize \textbf{Note:} RC = Random Calibration, Y8m = YOLOv8m
\end{table}

Table~\ref{tab:seg_accuracy} reports segmentation performance across configurations, isolating the effects of platform transfer, quantization, bias correction, and AVIS. Results are reported on a held-out test set representative of the deployment domain.

\paragraph{Platform transfer}
FP32 results across CPU and GPU show negligible variation, indicating that segmentation accuracy is primarily sensitive to quantization rather than execution backend.

\paragraph{Quantization impact}
INT8 with random calibration incurs a 6.6\% relative mAP reduction
(0.749 vs.\ 0.802), concentrated in mask boundary precision and small-object
completeness, consistent with activation range misestimation under
uninformative calibration.

\paragraph{Bias correction}
Adding bias correction improves mAP to 0.768 (+2.5\%), indicating that systematic layer-wise activation shifts contribute measurably to INT8 accuracy degradation and can be partially compensated.

\paragraph{AVIS + bias correction}
The combined strategy achieves the highest INT8 performance (mAP = 0.786, IoU = 0.767), recovering 69.8\% of the accuracy loss observed under random calibration relative to the FP32 baseline. This improvement is attributed to two complementary effects: AVIS improves activation range estimation via informative sample selection, while bias correction compensates residual distributional shifts in layer-wise activations.

\subsection{Qualitative Analysis}
\label{sec:qual}

\begin{figure*}[h]
    \centering
    \includegraphics[width=0.8\textwidth,keepaspectratio]
    {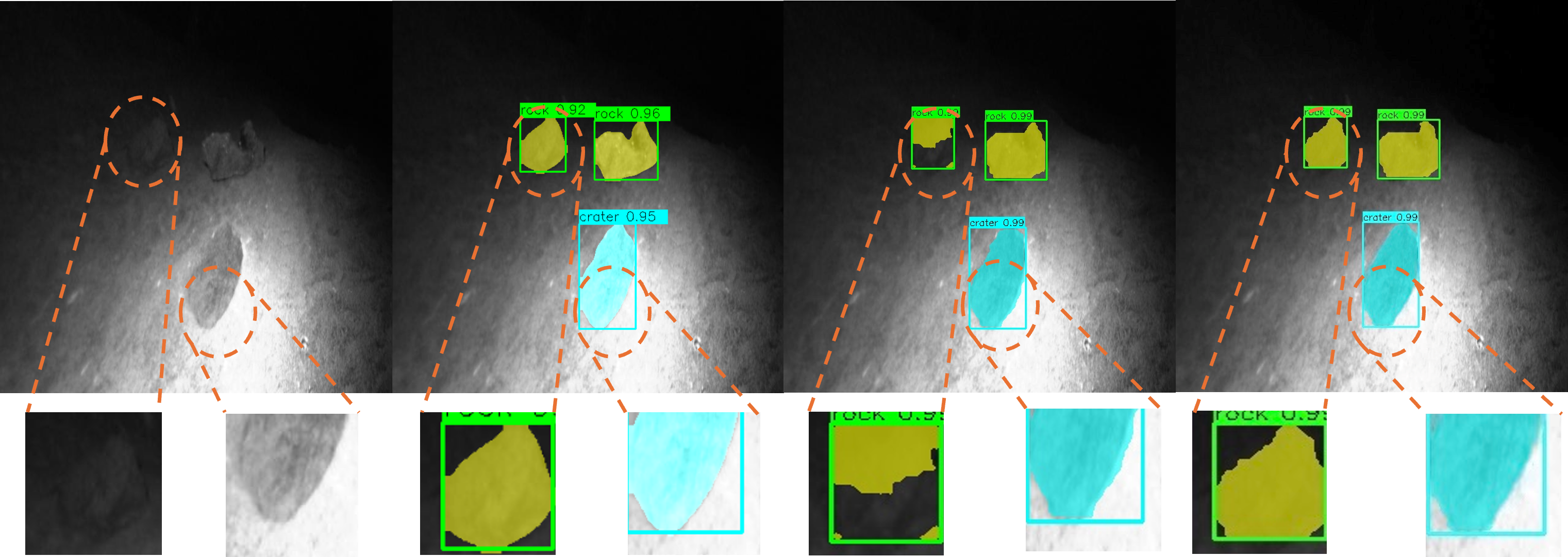}
    \caption{Instance segmentation under low-light lunar conditions.
    Left to right: input, FP32 reference, INT8 with random
    calibration (mask fragmentation and boundary distortion), INT8
    with AVIS + bias correction producing outputs closer to FP32 reference-level structure.
    Bottom: zoomed boundary crops.}
    \label{fig:results}
\end{figure*}

Figure~\ref{fig:results} compares segmentation outputs under representative low-light conditions. INT8 with random calibration produces mask fragmentation and boundary distortion due to activation clipping in fine-detail layers. AVIS + bias correction improves mask completeness and geometric fidelity, producing outputs closer to FP32-level structure, consistent with the quantitative trends in Table~\ref{tab:seg_accuracy}.

\subsection{Inference Efficiency}
\label{sec:inf_eff}

\begin{table}[t]
\centering
\caption{End-to-end inference latency, power, and energy per frame.}
\label{tab:deployment_efficiency}
\footnotesize
\setlength{\tabcolsep}{4pt}
\renewcommand{\arraystretch}{1.1}
\begin{tabular}{lccc}
\toprule
\textbf{Platform} & \textbf{Latency [ms]} & \textbf{Power [W]} & \textbf{Energy [J]} \\
\midrule
GPU     & 121 & 11.8 & 1.43 \\
CPU     & 537 &  8.3 & 4.46 \\
CPU-DPU & 309 &  5.7 & 1.76 \\
\bottomrule
\end{tabular}
\end{table}
The GPU achieves the lowest latency (121\,ms) but its 11.8\,W power draw exceeds the LuNiS Rover compute budget, making it unsuitable for typical lunar micro-rover power budgets. The CPU-only baseline exhibits the highest latency (537\,ms) and energy (4.46\,J/frame). The hybrid CPU-DPU configuration achieves 309\,ms at 5.7\,W (1.76\,J/frame), corresponding to a 60\% energy reduction versus CPU-only. Under the X Rover operational profile (0.1\,m/s velocity, 5\,s capture interval, 0.5\,m per frame), this provides more than 16$\times$ temporal margin for hazard detection within the perception-action cycle. All reported latencies include full on-device execution, covering DPU inference and CPU-based post-processing, with no offloading.

\subsection{Criticality Analysis}
\label{sec:crit_results}

Function-wise criticality analysis was performed following
Section~\ref{sec:criticality}, using \textit{vaitrace},
\textit{pmap}, \textit{massif}, and \textit{callgrind} for profiling.
The DPU-to-CPU execution time ratio is computed as the ratio of DPU kernel execution time to CPU postprocessing time, measured using \textit{vaitrace} and averaged over the evaluation set, yielding 0.2644. The DPU executes as a monolithic kernel and is modeled as a single functional entity.
\begin{figure}[h]
    \centering
    \includegraphics[width=0.8\columnwidth]{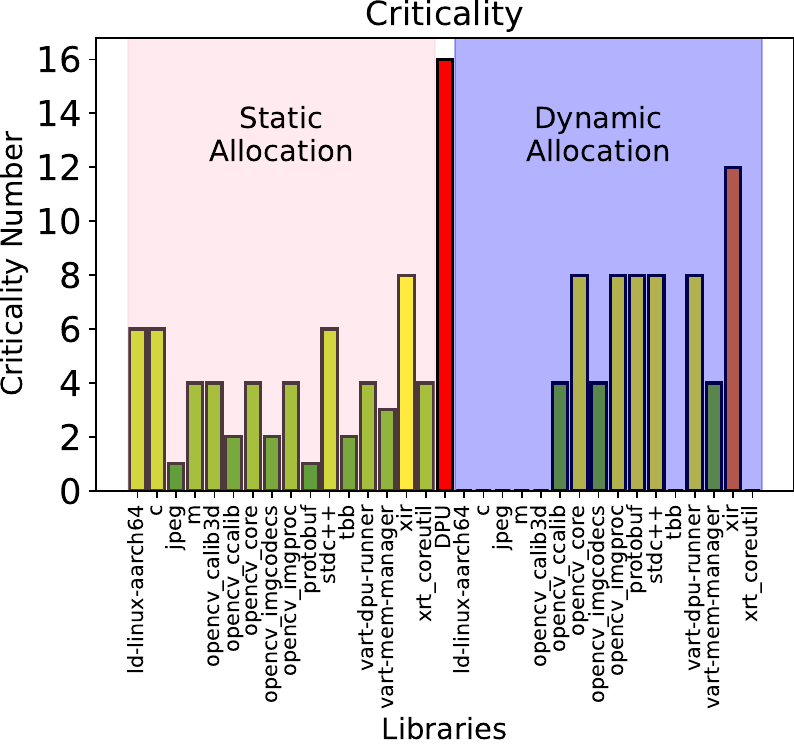}
    \caption{Per-library criticality under static and dynamic
    allocation. The DPU kernel exhibits the highest estimated criticality due to memory footprint and modeled propagation sensitivity.}
    \label{fig:criticality_results}
\end{figure}
\vspace{-10pt}
\begin{table}[h]
\centering
\caption{Per-block occurrence($O$) and severity($S$) ratings
(1--4 scale; 4\,=\,highest).}
\label{tab:severity_occurence}
\footnotesize
\setlength{\tabcolsep}{4pt}
\begin{tabular}{lcccc}
\toprule
 & \multicolumn{2}{c}{$O$} & \multicolumn{2}{c}{$S$} \\
\cmidrule(lr){2-3}\cmidrule(lr){4-5}
\textbf{Block} & Stat. & Dyn. & Stat. & Dyn. \\
\midrule
DPU Kernel         & 4 & 0 & 4 & 4 \\
C/C++ Backbone     & 2 & 0 & 3 & 4 \\
OpenCV             & 2 & 2 & 2 & 4 \\
Xilinx DPU Runtime & 2 & 3 & 4 & 4 \\
Math Libraries     & 2 & 0 & 2 & 4 \\
\bottomrule
\end{tabular}
\end{table}

\paragraph{Criticality distribution}
Figure~\ref{fig:criticality_results} and Table~\ref{tab:severity_occurence} show the per-block criticality distribution. The DPU kernel contributes most to the estimated system-level criticality due to its memory footprint and modeled propagation sensitivity, followed by the XIR runtime due to semantic fault sensitivity, while OpenCV and math libraries exhibit lower impact due to more localized fault propagation. Severity assignment follows a discrete ECSS-inspired ordinal scale, where 1 denotes negligible impact and 4 represents system-level or non-recoverable failure modes under radiation-induced faults. Dynamic allocations are assigned higher severity values due to reduced observability and recovery complexity under heap corruption, whereas the DPU kernel is assigned the maximum static severity ($S=4$) due to its dominant propagation potential.

\begin{table}[h]
\centering
\caption{Cumulative mitigation effect on global criticality.
Each row includes all preceding mitigations.}
\label{tab:mitigation}
\footnotesize
\begin{tabular}{lcc}
\toprule
\textbf{Case} & \textbf{$C$} & $\Delta$ \\
\midrule
Unmitigated                & 0.3389 & ---       \\
+ Static-only allocation   & 0.3161 & $-6.7\%$  \\
+ Selective TMR            & 0.2610 & $-17.4\%$ \\
+ EDAC \& memory scrubbing & 0.2316 & $-11.3\%$ \\
\midrule
\textbf{Cumulative}        &        & $\mathbf{-31.7\%}$ \\
\bottomrule
\end{tabular}

\vspace{2pt}
\footnotesize \textbf{Note:} EDAC\,=\,Error Detection and Correction,
$\Delta$ = relative to the preceding row.
\end{table}

\begin{figure*}[h]
    \centering
    \includegraphics[width=0.75\textwidth,keepaspectratio]
    {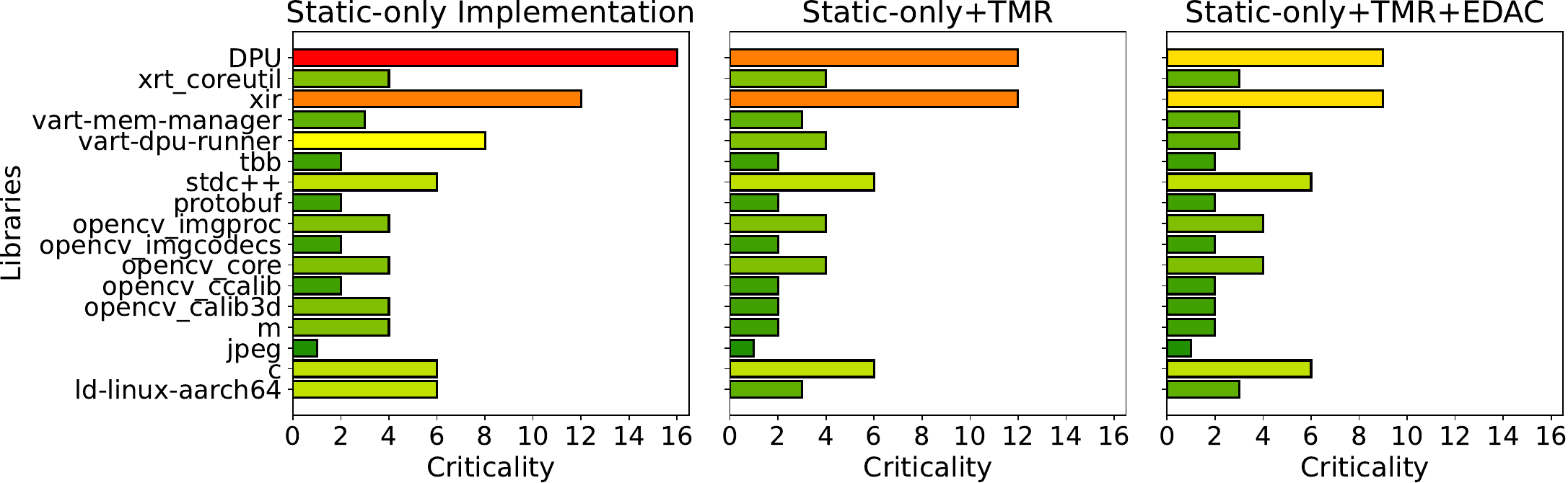}
    \caption{Cumulative criticality reduction under progressive
    mitigation. The DPU kernel and XIR runtime exhibit the largest
    reductions, validating the effectiveness of function-wise prioritization.}
    \label{fig:criticality_mitigation}
\end{figure*}

\paragraph{Progressive mitigation}
Progressive mitigation is evaluated in Table~\ref{tab:mitigation}, while Figure~\ref{fig:criticality_mitigation} visualizes the cumulative reduction in system-level criticality across successive mitigation stages. Static allocation reduces baseline criticality by 6.7\% by eliminating heap-associated risk. Selective TMR further reduces system vulnerability by 17.4\% by protecting high-impact compute blocks, while EDAC and memory scrubbing contribute an additional 11.3\% reduction by mitigating transient memory faults. Overall, these staged interventions reduce global criticality from 0.3389 to 0.2316, corresponding to a total reduction of 31.7\% relative to the unmitigated baseline.

AVIS and criticality-driven mitigation address complementary robustness dimensions. AVIS reduces sensitivity to quantization-induced perturbations in activation distributions, while criticality-driven mitigation reduces the estimated impact of radiation-induced faults at the system level under the adopted model.

\section{Conclusion}
\label{sec:conclusion}

This work presents a deployment-oriented framework for instance segmentation on resource-constrained lunar platforms. By integrating calibration-aware quantization, deterministic hardware execution, and function-wise fault analysis, the approach improves both inference reliability and system robustness. AVIS with bias correction recovers 69.8\% of quantization-induced accuracy loss (mAP 0.786 vs. 0.749 random calibration and 0.802 FP32), while enabling deterministic DPU execution at 309 ms and 5.7 W. Function-wise criticality
analysis identifies the DPU kernel and Xilinx runtime as dominant contributors to system-level risk, and targeted
mitigation—static allocation, selective TMR, and EDAC—reduced global
criticality by 31.7\%. These mechanisms are complementary: AVIS reduces quantization sensitivity by tightening activation ranges, while criticality-driven hardening mitigates radiation-induced faults that exceed model tolerance. The framework satisfies four key deployment requirements: quantization fidelity, deterministic execution, data-efficient calibration, and radiation-aware robustness. These results demonstrate that calibration-aware quantization and function-level analysis can significantly improve the reliability of AI perception under the extreme constraints of space robotics, where safety-critical autonomy is essential for rover operation.

\paragraph{Limitations}
The criticality analysis relies on analytical fault modeling rather than
physical fault injection; empirical validation under proton or heavy-ion
irradiation remains necessary. AVIS assumes a representative offline dataset; performance under distributional shifts (e.g., unfamiliar terrain types not included in calibration) remains unexplored. The 309\,ms latency is sufficient for
the current 0.1\,m/s profile but would require further optimization for faster
traversal or temporal fusion.

\paragraph{Future work}
Key directions include adaptive quantization scales responsive to accumulated
radiation dose, lightweight uncertainty estimation for per-prediction
confidence bounds, and multi-sensor fusion for landing-phase
redundancy. Formal bit-flip injection campaigns comparing AVIS-calibrated and
randomly calibrated models would empirically validate the hypothesized
relationship between activation range tightness and radiation fault tolerance.


\section*{ACKNOWLEDGMENT}
This research was done in the SAMLER-KI project, funded by the German Federal Ministry for Economic Affairs and Energy (BMWE, grant number 50RA2203A).


\bibliographystyle{IEEEtranS}
\bibliography{IROS2}


\end{document}